\documentclass{svproc}
\usepackage{url}

\usepackage{graphicx}  
\graphicspath{{./pics/}} 
\DeclareGraphicsExtensions{%
    .png,.PNG,%
    .pdf,.PDF,%
    .jpg,.mps,.jpeg,.jbig2,.jb2,.JPG,.JPEG,.JBIG2,.JB2}
    
\usepackage[caption=false]{subfig}

\usepackage{color}

\newcommand{\exTODO}[1]{}

\usepackage{bookmark} 

\begin{document}
\mainmatter              
\title{%
Haptics of Screwing and Unscrewing for its Application in Smart Factories for Disassembly}

\author{Dima Mironov\inst{1,2} \and
Miguel Altamirano\inst{1} \and
Hasan Zabihifar\inst{1} \and
Alina Liviniuk\inst{1} \and
Viktor Liviniuk\inst{1} \and
Dzmitry Tsetserukou\inst{1}}

\authorrunning{Dima Mironov et al.} 

\tocauthor{Dima Mironov,
Miguel Altamirano,
Hasan Zabihifar,
Alina Liviniuk,
Viktor Liviniuk, and
Dzmitry Tsetserukou}
\institute{Skolkovo Institute of Science and Technology (Skoltech), 127055 Moscow, Russia 
\and
\email{dima.mironov@skoltech.ru}}

\maketitle              

\thispagestyle{empty}
\pagestyle{empty}

\begin{abstract}
Reconstruction of skilled humans sensation and control system often leads to a development of robust control for the robots. We are developing an unscrewing robot for the automated disassembly which requires a comprehensive control system, but unscrewing experiments with robots are often limited to several conditions. On the contrary, humans typically have a broad range of screwing experiences and sensations throughout their lives, and we conducted an experiment to find these haptic patterns. Results show that people apply axial force to the screws to avoid screwdriver slippage (cam-outs), which is one of the key problems during screwing and unscrewing, and this axial force is proportional to the torque which is required for screwing. We have found that type of the screw head influences the amount of axial force applied. Using this knowledge an unscrewing robot for the smart disassembly factory RecyBot is developed, and experiments confirm the optimality of the strategy, used by humans. Finally, a methodology for robust unscrewing algorithm design is presented as a generalization of the findings. It can seriously speed up the development of the screwing and unscrewing robots and tools.
\end{abstract}

\section{Introduction}

The e-waste recycling is a subject in which interest increases rapidly due to the growing importance of energy conservation, material resources, and landfill capacity. For example, Apple corporation recently developed a disassembly line for iPhone 6 \cite{liam}. The goal of the project RecyBot is to develop a universal high-speed intelligent robotic system for electronics recycling. RecyBot consists of several robots, each tailored to perform a specific task, whose joined target is to disassemble smartphones at the component level and enable material recovery. Previously published attempts of the automation of mobile phone disassembly has been primarily focused on the disassembly system design \cite{kopacek_robotized_2003,kopacek_intelligent_2006} and are outdated due to the recent advantages in computer vision techniques, which question the assumptions of the possible level of the system autonomy \cite{vongbunyong_basic_2013}. Therefore, RecyBot focuses on the development of the series of smart robots with computer vision, which automate operations, which are often performed during electronic waste disassembly.

The unscrewing operation is an essential part of the RecyBot project because screws are a central category of fasteners and are found in the majority of electronic products \cite{disassembly_book}. The screw removal has to be performed via non-destructive means, since it often precedes the battery removal, which, if damaged, may cause more ecological damage, than the ecological benefits of recycling. The previous attempt to build an unscrewing robot was described by Chen et al. \cite{chen_robot_2014}, who built a robot to assist humans in electric vehicles batteries disassembly. They use compliant trajectories, taught by humans for actual unscrewing, who can also assist the robot in case of failure, but the focus in the paper is on the tool change procedure, and it omits the details of the unscrewing itself. An extensive analysis and detection of common errors during automatic unscrewing are done by Apley et al. \cite{apley_diagnostics_1998}. They use a screwdriver, attached to a DC motor with a potentiometer to read the screwdriver head rotation angle, and continuously estimate the torque from the motor current. They successfully detect normal unscrewing, cam-out (slippage between the screwdriver and the screw), the situation, when the screwdriver missed the screw, and the situation, when their system cannot provide enough torque. However, they do not report, how to overcome these problems.

\begin{figure}[!t]
    \centering
    \includegraphics[width=\linewidth]{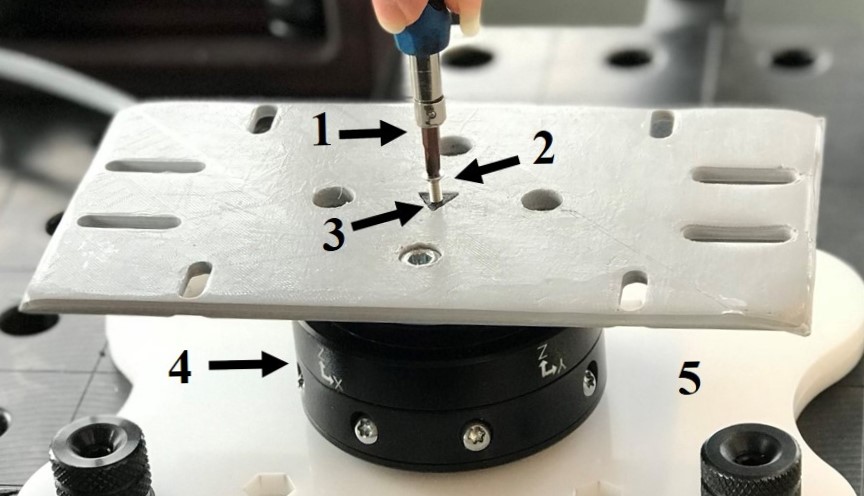}%
    \caption{The experimental setup. The disposable plastic prism (2) with a 3 mm hole for the M3 screw is fixed inside the prism holder (3). The prism holder is attached to a 6 DOF force and torque sensor (4), which is fixed to a heavy table using an acrylic fixture (5). The screw is unscrewed by a participant with a corresponding screwdriver (1). The center line of the screw is positioned along the Z-axis, which points up. Force and torque along Z-axis are recorded with the frequency of 100 Hz.}
    \label{fig:experimental_setup}
\end{figure}

The cam-out is an important problem during osteosynthesis procedures in medicine. In such critical applications, it is crucial to teach humans in safe conditions. Majewicz et al. built a simulator to teach surgeons to screw, using the torque data gathered by an automated screwdriver and validated the teaching with the help of more experienced surgeons \cite{majewicz_design_2010}. The cam-out in the same context was studied in \cite{behring_slippage_2002}, where screws were intentionally damaged, and this damaging torque was measured, but the paper does not describe a way to prevent slippage which occurs not from metal deformations, but from the screwdriver coming out of the screw.

This paper presents a methodology for robust robotic unscrewing development, using skilled humans data. The following is structured in two major parts: an experiment on human screwing and unscrewing is presented in Section \ref{sec:Human Experiment} and a description of the application of its results to an unscrewing robot is given in Section \ref{sec:Unscrewing Robot}. First, experimental setup and procedure are described in Section \ref{sec:methods}, where it is measured how humans perform screwing and unscrewing operations in different conditions. It is assumed that screwing and unscrewing patterns of humans, who incorporate their broad experiences in the patterns which they use, may lead us to insights into robotic screwing and unscrewing. Two critical characteristics of the process have been measured: force applied along the screwdriver and screwing or unscrewing torque. Then it is observed how do humans prevent cam-out (screwdriver slippage), which is one of the most common failures \cite{apley_diagnostics_1998}, and how do they decide, when to stop. The conclusions from the human experiment are presented in Section \ref{sec:results}. 

In the second part, an unscrewing robot is described, and the results of the first part are applied. First, mechanical design is described in Section \ref{sec:Gripper Design}, then the control system, which is inspired by the experiment from the first part, is elaborated on in Section \ref{sec:Unscrewing Control}. The experiments with the robot are further described in Section \ref{sec:Robot Unscrewing Results}. The overall approach is discussed and Summarized in Section \ref{sec:conclusions}.

\section{Human Experiment}
\label{sec:Human Experiment}

The principal approach of this paper is to design robot control algorithms based on the coefficients which are obtained from humans. Screwing and unscrewing are rather complex procedures, and humans integrate their past experiences into their behavior and pattern. Human screwing and unscrewing are classified, their important characteristics are measured, and then these parameters are transferred to robotic unscrewing. The experiment is designed to get a thorough and broad view of the patterns human use.

\subsection{Experimental Design}
\label{sec:methods}

\begin{figure}[!th]
  \centering%
  \subfloat[Disposable plastic prism with a $3\ mm$ pilot hole.]{%
         \label{fig:prism_hole}%
         \includegraphics[width=0.2\linewidth]{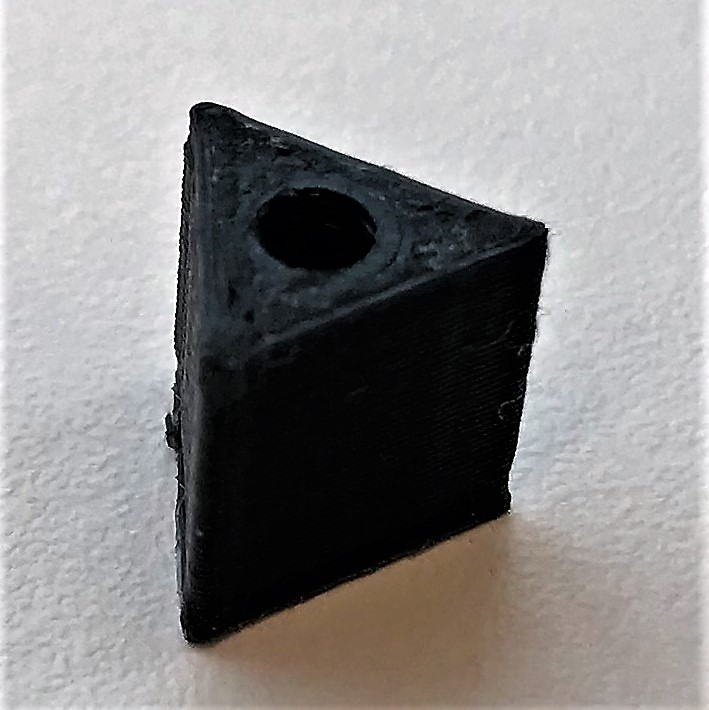}%
} \qquad
\subfloat[Disposable plastic prism with an M3 nut added during the 3D printing.]{%
         \label{fig:prism_nut}%
         \includegraphics[width=0.2\linewidth]{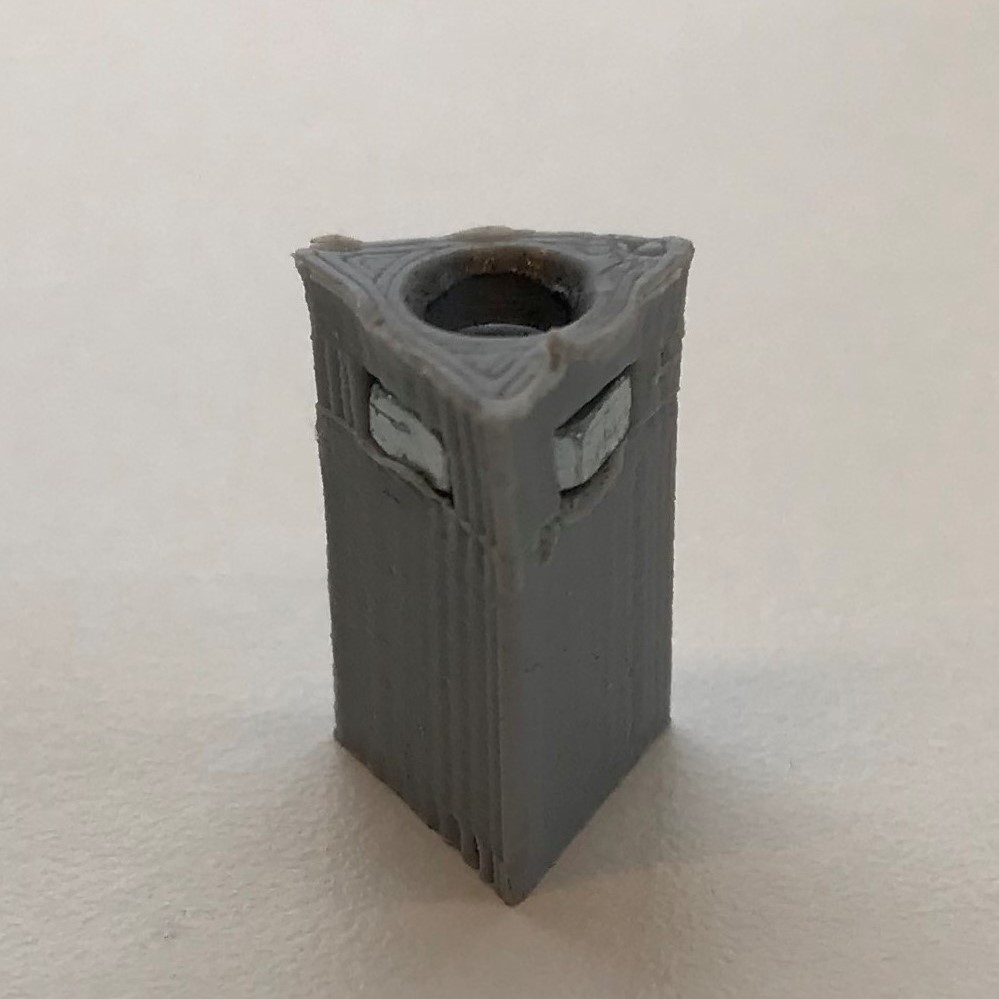}%
} \qquad
\subfloat[Phillips screw head]{%
         \label{fig:phillips_screw}%
         \includegraphics[width=0.2\linewidth]{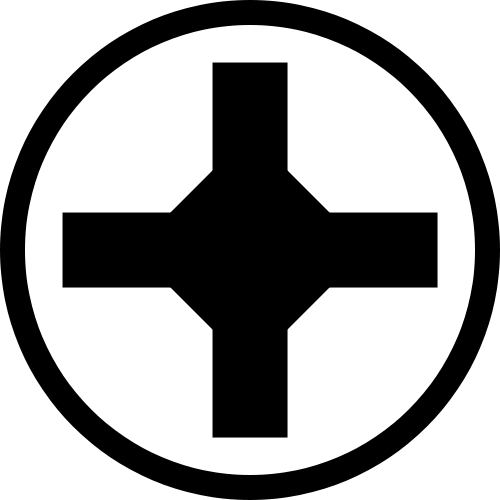}%
} \qquad
\subfloat[Internal Hex screw head, also called Allen screw]{%
      \label{fig:hex_screw}%
      \includegraphics[width=0.2\linewidth]{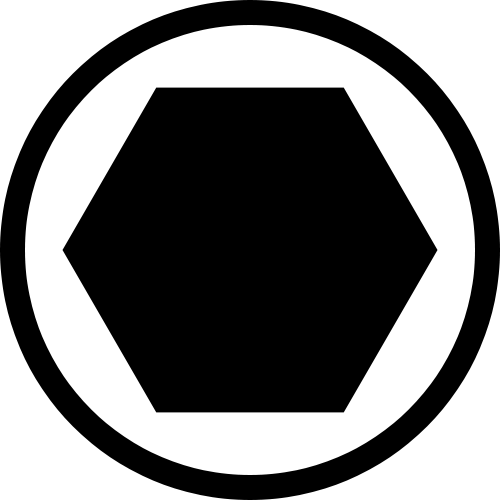}%
    }

  \caption{Disposable plastic prisms, where the participants screw and unscrew. Two types of screws were used, Phillips and Internal Hex.}
   \label{fig:prism_screwheads}
\end{figure}

The principal factors to measure and control in the screwing and unscrewing operation are the force and the torque which are applied perpendicular to the direction of the surface and along the axis of the screw (axial force). However, the torque applied during the interaction with the real smartphone screws was too low to measure with the available sensors. To measure the human patterns of screwing and unscrewing, an experimental setup has been built as shown in Fig. \ref{fig:experimental_setup}. A plastic holder (3) was designed to keep the pilot holes and nuts object, where the screwing process happened. The object, where the participants screwed and unscrewed, was designed as a disposable triangular prism (2) with a pilot hole in the center of it of $3\ mm$ diameter. These prisms were fixed in the center of the holder and replaced after being used. To reproduce the presence of a nut in some smartphones, an M3 nut was added to some prisms during the 3D printing $5\ mm$ below the end of the prism. During all the experiment were used M3 x $8\ mm$ screws with two different heads: Phillips and internal Hex, with their corresponding screwdriver (1). In the Fig. \ref{fig:prism_screwheads} both the disposable plastic prisms and the heads of the two different screws used during the experiment are shown.

The holder was attached to a Robotiq 6 DOF force and torque sensor FT300 (4) and the pilot hole in the prism were aligned with a sensor Z-axis. This sensor was chosen because of its frequency of $100\ Hz$ for data output and the low noise signal of $0.1\ N$ and $0.003\ N \cdot m$ in $F_z$ and in $M_z$ respectively \cite{robotiq_ft300}, that allowed getting enough data for the purposes of this experiment. The sensor was fixed to a massive and stiff table (Siegmund Professional S4 welding table) using an acrylic base (5). 

Ten participants, who used screwdrivers at least occasionally for their duties, were recruited for the experiment, seven men, and three women, in an age range from 22 to 40 years. They were asked to screw the screw in the prism, holding the screwdriver only with one hand during all the process, and, if it was necessary, to hold the screw with the other hand. After that, they were asked to unscrew the same screw. The experiment consisted of 9 screwing-unscrewing operations in 6 different conditions for each of the participants: 2 times with a new Phillips screw and 1 time with an Allen screw vertically; 1 time with a Phillips screw, and 1 with an Allen screw horizontally; 3 times with a Phillips screw, but in the prism with the nut vertically; and one time with a Phillips screw, but with a screwdriver that did not correspond with the screw (we call this condition "exceptional" and other conditions "typical"). The disposable prism has been replaced after each unscrewing, to avoid the influence of the thread, which appeared in the plastic. Some of the participants were also asked to screw and unscrew the screw into the disposable plastic prism without applying any force in the direction of the screw.

\subsection{Experimental Results}
\label{sec:results}

\begin{figure}[!thb]
  \centering%
  \subfloat[Typical pattern of absolute values of force along the screw axis and torque during unscrewing of a Phillips screw from a disposable plastic prism. The black line represents the measurements. Green dashed line is the fit of the local maximums with a spline. It represents the probable pattern, suitable for a robot in the same task. ]{%
         \label{fig:typical_unscrewing}%
         \includegraphics[width=0.46\linewidth]{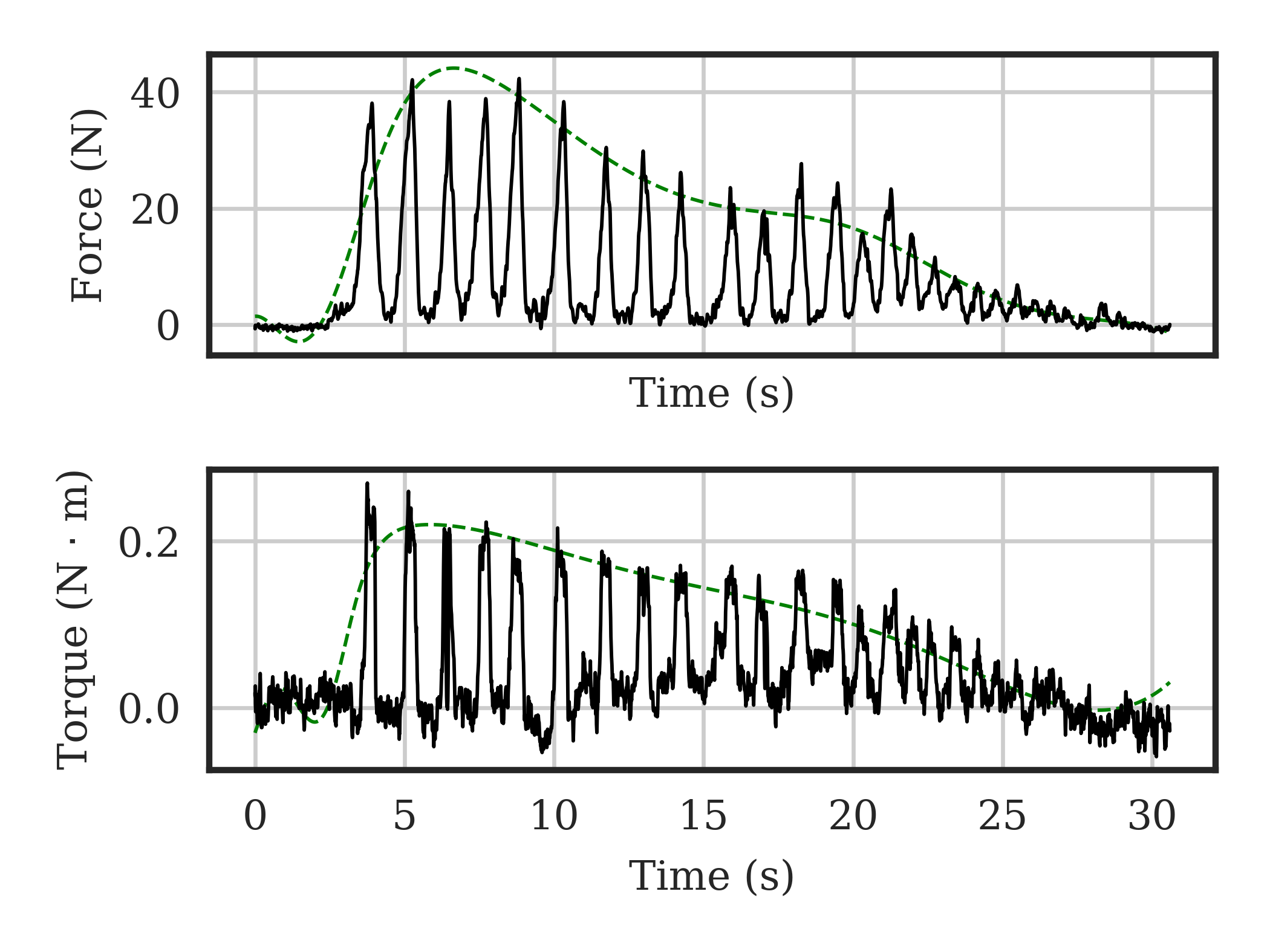}%
} \qquad
\subfloat[Relationship between force and torque for the time series from (a). Points represent individual measurements and the red line is the least squares approximation.]{%
      \label{fig:typical_force_torque}%
      \includegraphics[width=0.46\linewidth]{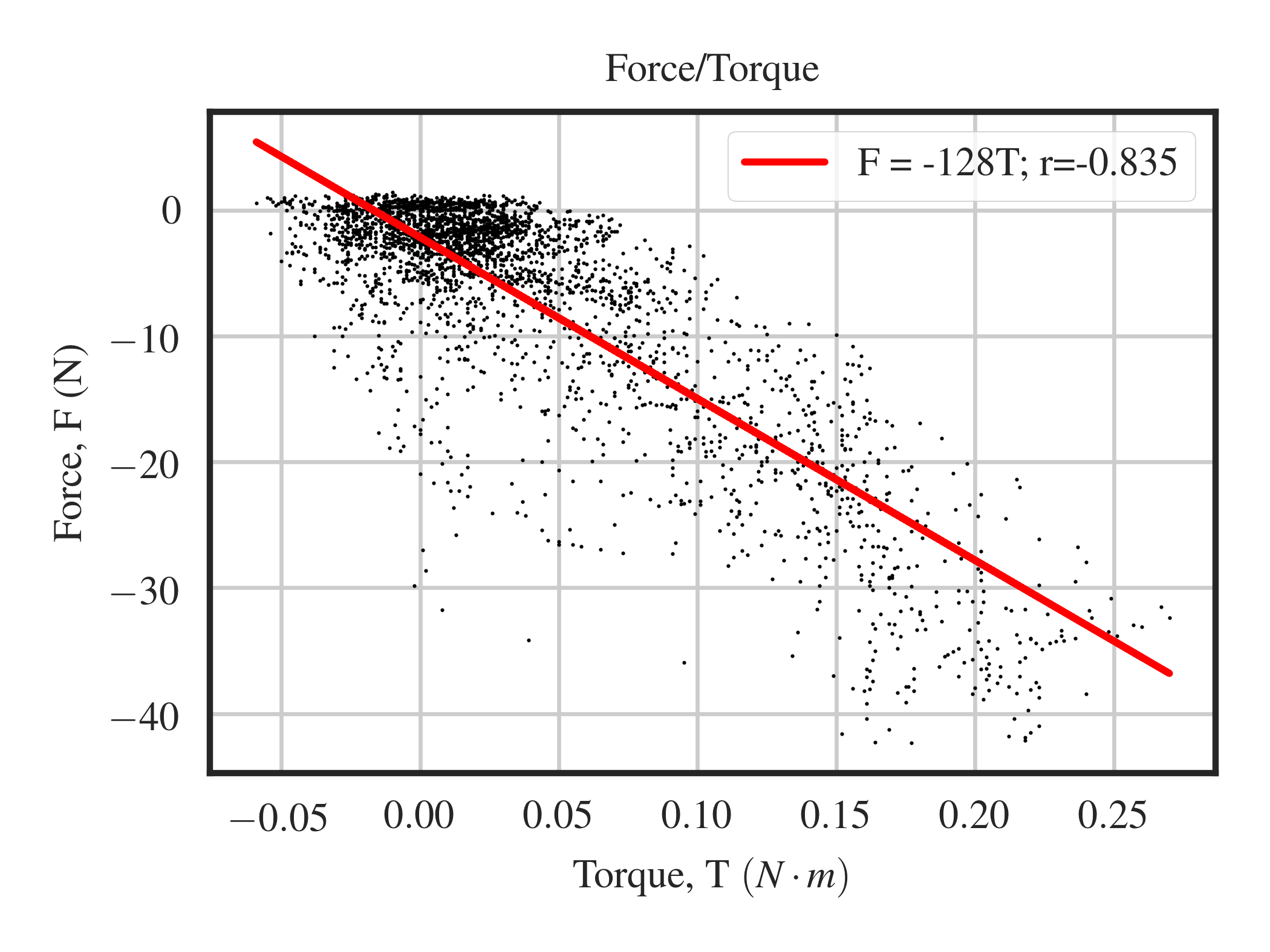}%
    }
    
  \subfloat[Typical pattern of absolute values of torque during screwing of a Phillips screw into a disposable plastic prism. The black line represents the measurements. Green dashed line is the fit of the local maximums with a spline. It represents the probable pattern, suitable for a robot in the same task.]{%
         \label{fig:typical_screwing}%
         \includegraphics[width=0.9\linewidth]{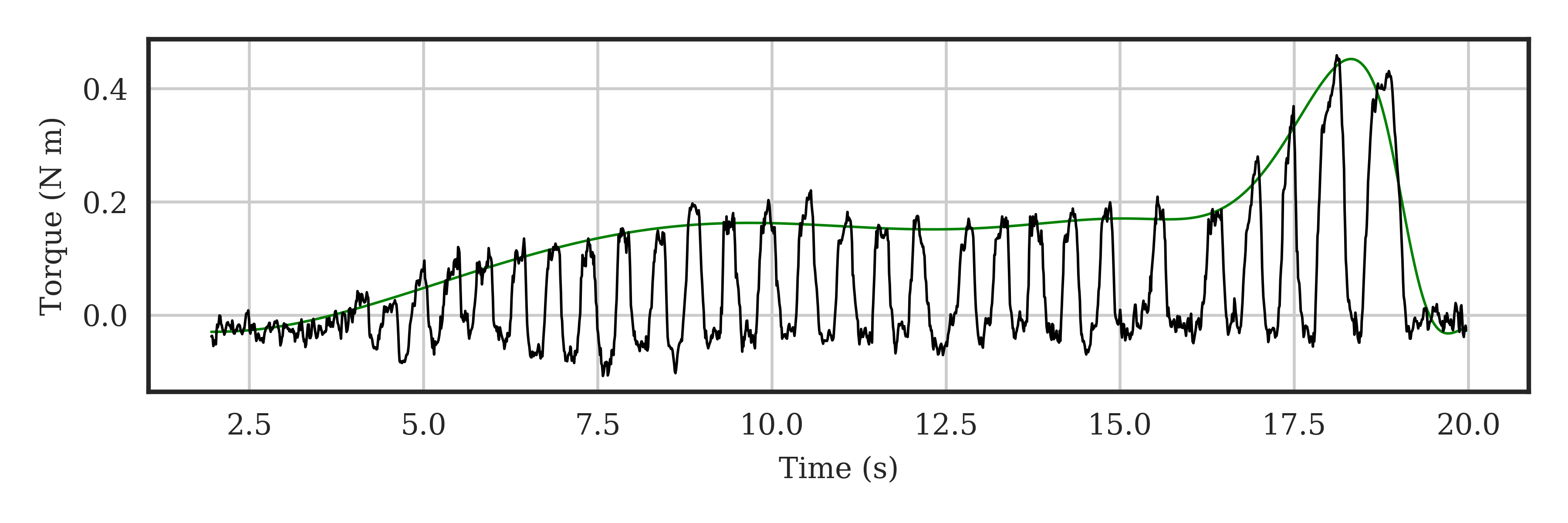}%
}

  \caption{Typical patterns of force along the screw axis and torque during screwing and unscrewing an M3 Phillips screw in the disposable plastic prism.}
   \label{fig:typical_patterns}
\end{figure}

The typical patterns of force and torque during unscrewing of Phillips screws in plastic disposable prism are shown in Fig. \ref{fig:typical_patterns}. The oscillatory pattern with the frequency of $(1.3\ Hz \pm 40\%)$ is observed in all the measurements from all the participants both in torque and in force and is independent of age or gender. It is caused by the need for humans to regrasp the screwdriver due to lack of joints with limitless rotation.

The pattern of local maximums of torque and force represents the moments when real screwing and unscrewing were happening. The torque required to rotate the screw is defined by the friction, which is defined by the environment: it depends on the state of the thread in the plastic and does not depend on the rotation speed in a wide range of speeds (see Section \ref{sec:Robot Unscrewing Results}). During unscrewing from a disposable plastic prism (see Fig. \ref{fig:typical_unscrewing}), the maximal torque is applied at the beginning of the procedure, and then it decreases gradually, which is defined by the decrease of the length of the screw inside the plastic and thus the friction decrease. In Fig. \ref{fig:typical_screwing} a typical screwing pattern is shown. The torque is gradually increasing in the first seconds while the screw is dipping into the plastic. Then it remains constant indicating, that most friction is occurring due to the need to cut a new treading in the plastic. Final torque increase indicates the reach of the plastic surface with a screw head. Humans feel the torque increase and stop screwing.

A correlation between force and torque applied simultaneously is observed, which can be seen in Fig. \ref{fig:typical_force_torque}. For all of the measured conditions the absolute value of the correlation coefficient $r = (0.75\pm0.13)$, which indicates a significant relation. We than denote the Force/Torque ratio $\nu$ and take it as a characteristic of the screwing or unscrewing procedure in the specific conditions. Two reasons to apply force simultaneously with torque are, first, the need to avoid screw slippage and, second, the nature of human arm, which uses muscles to produce torque via forces \cite{tsetserukou2010}. To estimate the importance of the first factor, we compare the Phillips screws and screws with internal Hex. The average $\nu$ for screwing the Phillips screw operations is $\nu = (106\pm37)\ m^{-1}$ which is significantly higher than $\nu = (57\pm25)\ m^{-1}$ for screwing of the screws with internal Hex. Since the second factor does not depend on the screw, the difference is contributed to only by the slippage avoidance. Thus, the screws with internal hex are less affected by slippage, than the Phillips screws, being screwed or unscrewed with the same force.

\begin{figure}[!thb]
  \centering%
 \includegraphics[width=0.7\linewidth]{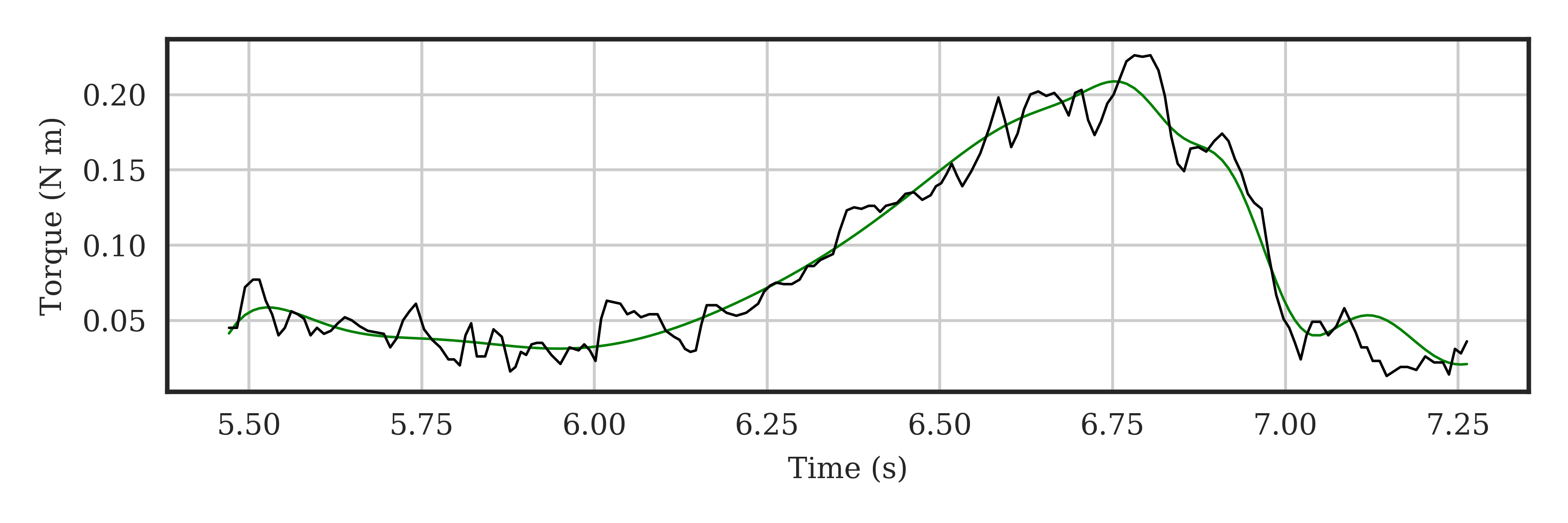}%
  \caption{Typical pattern of absolute values of torque during screwing of a Phillips screw into a disposable plastic prism with a nut. The black line represents the measurements. Green dashed line is the fit of the local maximums with a spline. It represents the probable pattern, suitable for a robot in the same task.}
   \label{fig:typical_nut}
\end{figure}

In Fig. \ref{fig:typical_nut}, a typical pattern of screwing a screw in the nut is shown. During the first seconds the torque is hardly distinguishable from noise, but when the screw head touches the surface, the torque increases, human feel it and stop screwing. The typical unscrewing pattern is very similar, but inverted in time and requires smaller maximum torque, than the maximum torque which was applied during screwing.

\begin{figure}[!thb]
  \centering%
 \includegraphics[width=\linewidth]{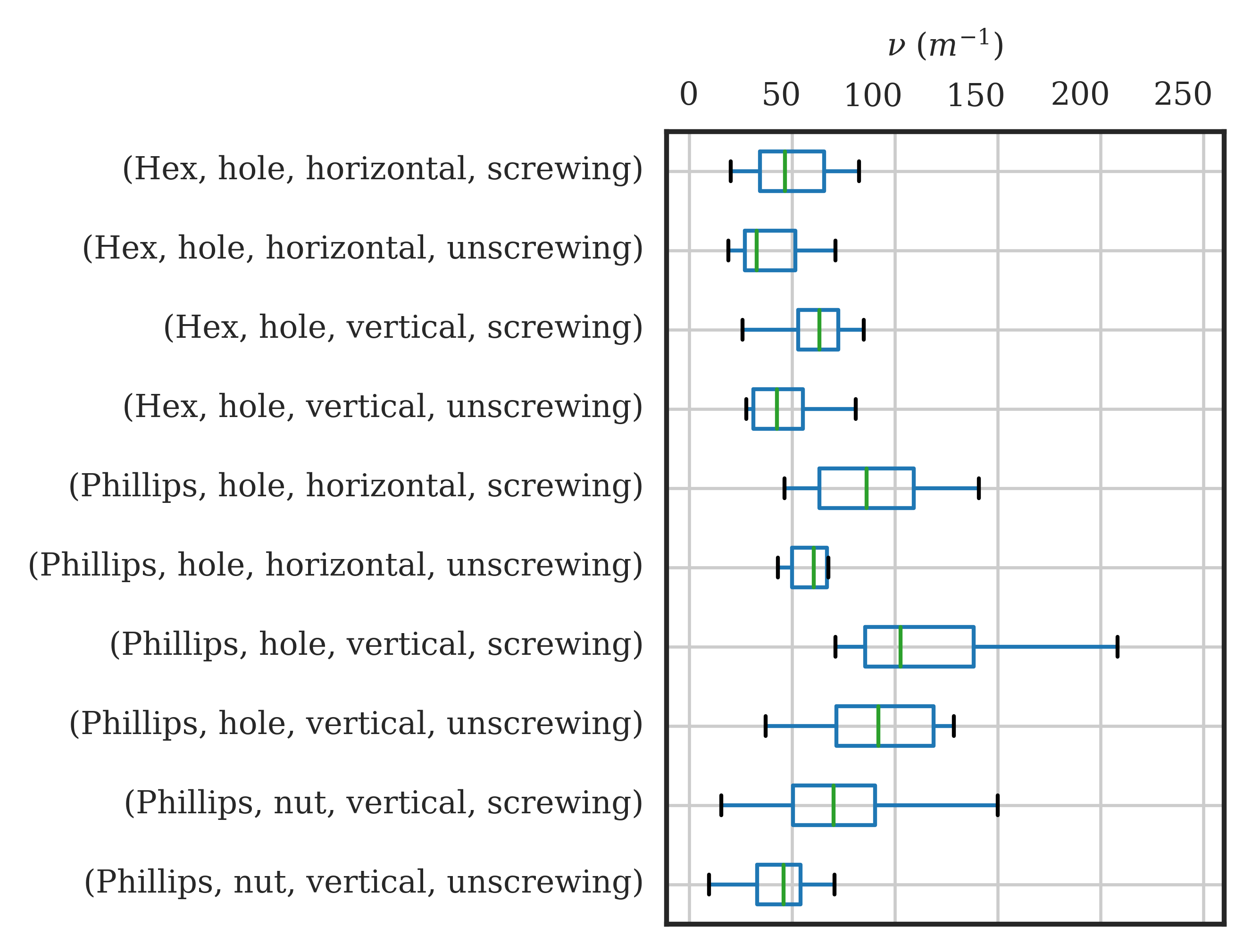}%
  \caption{Bar plot of Force/Torque ratio $\nu\ (m^{-1})$ in different conditions. The conditions in the brackets represent (the screw type, "hole" is for the disposable plastic prism and "nut" is for a prism with a nut, the orientation of the screw,
  screwing or unscrewing).}
   \label{fig:whiskers}
\end{figure}

In Fig. \ref{fig:whiskers} a box plot of the $\nu$ coefficient is shown for all the typical conditions. The $\nu$ for the exceptional condition when the screwdriver does not match the screw is much larger, in the order of $300 \pm 250\ m^{-1}$ and was omitted to keep the scale sensible. The large $\nu$ is reasonable, since it was very hard for the participants to apply any significant torque, because of continuous slippage. Moreover, during operations performed by a participant, when explicitly asked not to apply any force, and only apply torque, the slippage was also constantly occurring. This confirms the hypothesis, that the function of the typically applied force is to avoid slippage. The mean values of $\nu$ differ between screwing and unscrewing. For the taken conditions $\nu$ is typically in the range from 25 to 140.

The difference between horizontal and vertical operations could be introduced with the need to press the screwdriver in the screw in the horizontal position even during regrasp, to avoid its fall and caused by the relative comfort of force application in different orientations. However, the difference is not significant even for unscrewing Phillips screws. A p-value from Mann–Whitney U test for the two distributions is $p = 0.065 > 0.05$ and thus influence of the orientation cannot be confirmed by this study.

\subsection{Discussion}

The discovery of the linear relationship between applied torque and force is a key to designing a robust screwing or unscrewing system. Since the target torque is defined by the environment (see Section \ref{sec:Robot Unscrewing Results}), the robust algorithm should use force control and define the force from torque, using the coefficient, obtained from humans for the exact conditions.

\section{Unscrewing Robot}
\label{sec:Unscrewing Robot}

\begin{figure}[th]
    \centering
    \includegraphics[width=0.8\linewidth]{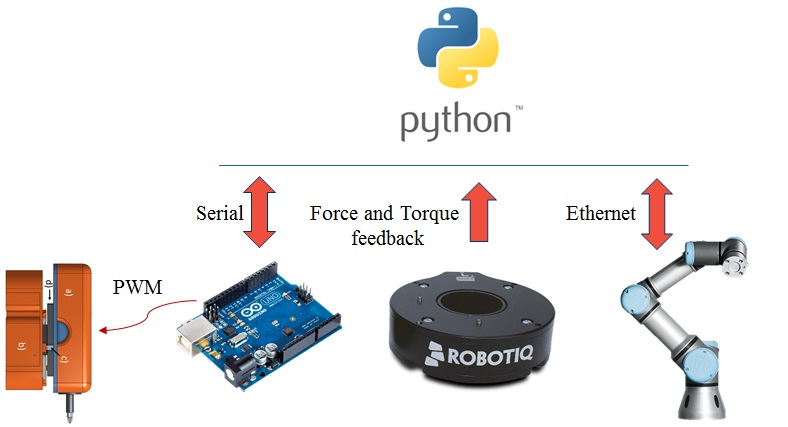}%
    \caption{The general scheme of communication in the testbed for robotic unscrewing.}
    \label{fig:system2}
\end{figure}

An unscrewing robot is being developed as a part of the RecyBot project. It is equipped with computer vision to detect screws and a gripper to unscrew them. To perform unscrewing well, the force applied to the screw has to be controlled. Analyzing data from human experiment helps to define appropriate pattern and thresholds for force and torque in robotic implementation.  Maximum force, maximum torque, and force-torque relation are the important parameters that are used. The setup diagram is shown in Fig. \ref{fig:system2}

\label{sec:unscrewing_robot}

\subsection {Mechanical Design}
\label{sec:Gripper Design}

A collaborative robot UR3 from Universal Robots was used to move the gripper. The design of the gripper, that integrates passive compliance, was done with the objective to increase the force feedback precision in the compliant direction of Z-axis \cite{ham_compliant_2009}. The passive compliance is implemented with a linear bearing, a spring system, and a linear potentiometer. The relative position of the potentiometer indicates the force applied to the screw as defined by the spring system. The linear potentiometer was calibrated using a Robotiq 6 DOF force and torque sensor FT150. The screwdriver in the gripper is attached to a DC motor from Maxon, the current from which is measured and used to determinate the torque applied during the operation. Both force and torque, measured by the internal gripper sources, are used in the real unscrewing, to avoid the usage of the expensive 6 DOF sensor in the setup.

\subsection{Screwing and Unscrewing With Slippage Avoidance}
\label{sec:Unscrewing Control}

\begin{figure}[!th]
  \centering
  \includegraphics[width=0.9\linewidth]{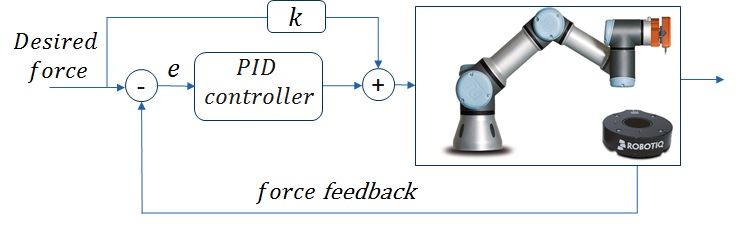}%
  \caption{Structure of the control scheme. Position is controlled using the force feedback.}
  \label{fig:controller2}%
\end{figure}

Unscrewing procedure is controlled using a force control algorithm. The PID force controller with feed-forward signal was designed and in Fig. \ref{fig:controller2} the structure of the control system is shown. The control input signal is a position command which is applied to the robot end-effector. The force feedback is received from the calibrated potentiometer.

In addition, to prevent cam-out, which is one of the most common failures \cite{apley_diagnostics_1998}, the force is applied according to the torque multiplied by the Force/Torque ratio $\nu$, which is obtained from human experiment data. When the robot senses the slippage of the screwdriver, the force is increased to prevent it. In fact, when the force, which is applied during the unscrewing is controlled, one can make the robot's activity closer to human activity as a clever actor.

\subsection{Robot Screwing and Unscrewing Results}
\label{sec:Robot Unscrewing Results}

\begin{figure}[!h]
  \centering
  \includegraphics[width=0.9\linewidth]{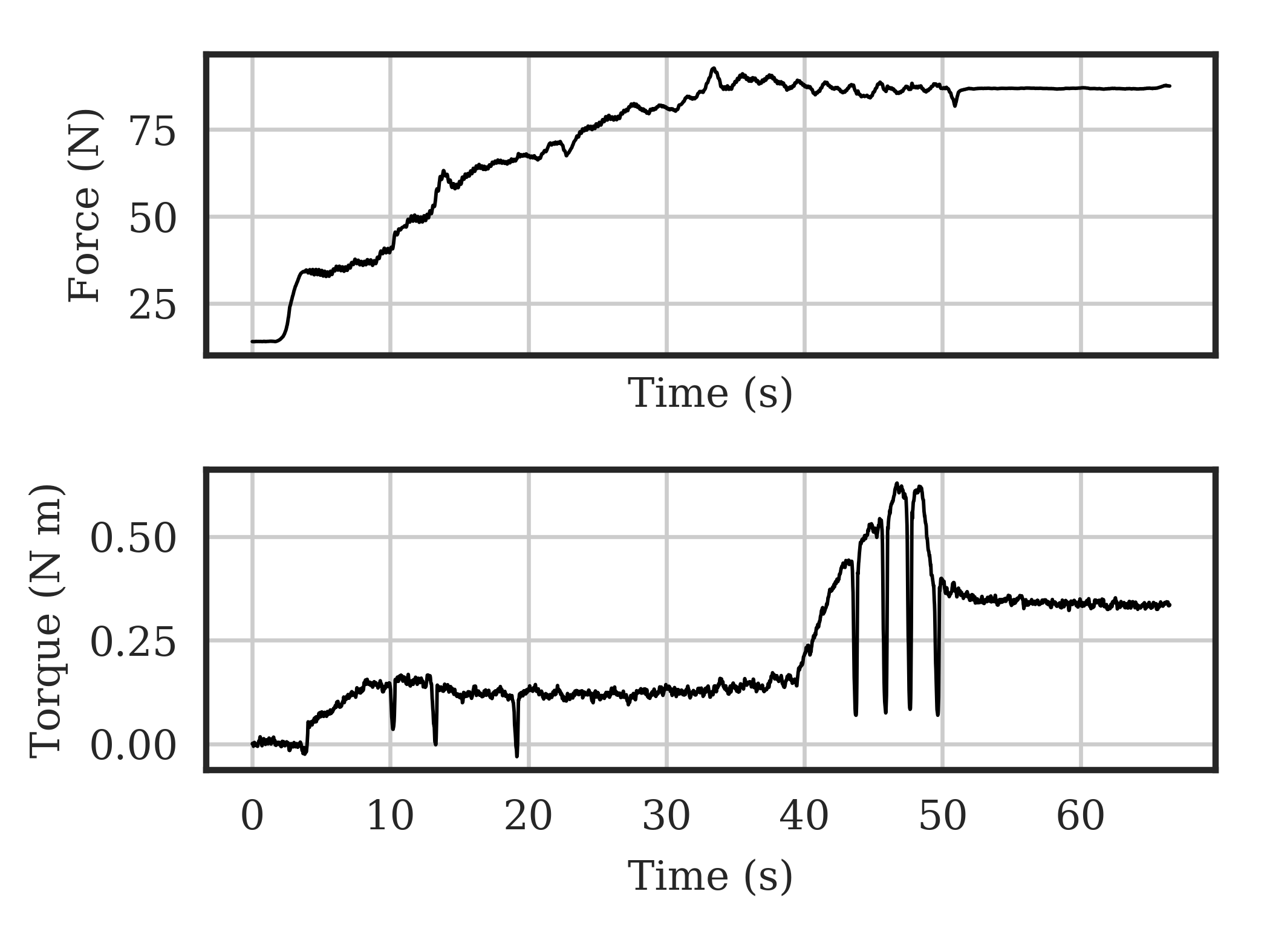}%
  \caption{Force and torque values obtained from a robot, screwing thew screw into a disposable plastic prism. The robot is set to gradually increase the force until the maximum value is reached. Cam-outs correspond to sharp drops in torque. Their frequency is decreasing from 5 s to 22 s when the force reaches the value twice higher than the one, computed from $\nu$, the Force/Torque ratio in humans.}
  \label{fig:robot_force_torque}%
\end{figure}

The robot was programmed with an algorithm, close to the proposed, in the demonstration purposes. In Fig. \ref{fig:robot_force_torque} the obtained force and torque are shown. The cam-outs are clearly visible as drops in the torque. Their detection was implemented using a threshold. The force was being gradually increased, with higher increase speed, when slippage was detected. The frequency of the cam-outs notably drops with the increase of the force from 5 to 25 s. This means sufficiently large force has been reached to avoid cam-outs. The head of the screw touched the plastic at 40 s. The torque started to grow rapidly. When the Force/Torque ratio $\nu$ lowered due to the increase in torque, the slippage started to occur again after 43 s. Then the thread is overturned since no torque limit has been implemented in the prototype algorithm.

To check the hypothesis, that the torque does not depend on the speed of the screw rotation, an experiment has been performed with a robot. An unscrewing speed was changed in a range from 22.5 to 360 degrees per second and the measured unscrewing torque in the same conditions was $T=(0.19 \pm 0.03) N\cdot m$. This confirms the hypothesis of the independence of the friction on the speed.
   
\subsection{Discussion}

The novel end-effector has been designed, which is able to screw or unscrew the screws with precise force feedback. The main feature of the robot is the desired Force/Torque relationship $\nu$ of the robot unscrewing, which is gathered from the human experiments. The results of the robotic unscrewing agree with the results of the human experiments and demonstrate the universality of the conditions of the successful unscrewing found in the previous part. The human experiment data helps us to find out the well-performing desired force pattern. 

\section{Discussion and Conclusions}
\label{sec:conclusions}

A typical frequency of human screwing and unscrewing is reported here to be $(1.3\ Hz \pm 40\%)$ without any dependence on age and gender. It seems to correlate more with the strength of the participants, but we did not measure this variable.

A significant correlation between applied torque and force during screwing and unscrewing is discovered in this paper, and it should be a reasonable assumption for any screwing and unscrewing procedures design. 

In this paper, a new approach for robotic screwing and unscrewing procedure has been proposed. To choose the constants for the robot screwing and unscrewing algorithm, one has to, first, measure the Force/Torque ratio $\nu$ humans typically apply in the exact conditions, and then implement the force control, based on the continually measured torque and obtained $\nu$. One can increase the coefficient to introduce a safe margin but should consider the fragility of the environment. Also, the maximum torque humans use has to be chosen as a threshold for screwing condition detection.

This methodology can seriously speed up the development of the screwing and unscrewing robots and tools, such as simulators for surgeons and other professionals. It allows direct transfer of haptic knowledge from the more knowledgeable ones to the novices. 

\section*{Acknowledgment}
We acknowledge all the volunteers, who participated in our experiments.

\bibliographystyle{styles/spmpsci}
\bibliography{IEEEabrv,SkolTech}

\begin{thebibliography}{10}
\providecommand{\url}[1]{{#1}}
\providecommand{\urlprefix}{URL }
\expandafter\ifx\csname urlstyle\endcsname\relax
  \providecommand{\doi}[1]{DOI~\discretionary{}{}{}#1}\else
  \providecommand{\doi}{DOI~\discretionary{}{}{}\begingroup
  \urlstyle{rm}\Url}\fi

\bibitem{robotiq_ft300}
Robotiq {FT}300 specifications.
\newblock
  \urlprefix\url{https://assets.robotiq.com/production/support_documents/document/specsheet-FT300-Nov-08-V3_20171116.pdf?_ga=2.252630985.1712588583.1517308561-1050351704.1509722697}

\bibitem{apley_diagnostics_1998}
Apley, D.W., Seliger, G., Voit, L., Shi, J.: Diagnostics in {{Disassembly
  Unscrewing Operations}}.
\newblock International Journal of Flexible Manufacturing Systems
  \textbf{10}(2), 111--128 (1998).
\newblock
  \urlprefix\url{https://link.springer.com/article/10.1023/A:1008089230047}

\bibitem{behring_slippage_2002}
Behring, J.K., Gjerdet, N.R., Mølster, A.: Slippage between screwdriver and
  bone screw.
\newblock Clin. Orthop. Relat. Res. (404), 368--372 (2002)

\bibitem{chen_robot_2014}
Chen, W.H., Wegener, K., Dietrich, F.: A robot assistant for unscrewing in
  hybrid human-robot disassembly.
\newblock In: Robotics and {{Biomimetics}} ({{ROBIO}}), 2014 {{IEEE
  International Conference}} On, pp. 536--541. {IEEE} (2014).
\newblock \urlprefix\url{http://ieeexplore.ieee.org/abstract/document/7090386/}

\bibitem{ham_compliant_2009}
Ham, R., Sugar, T., Vanderborght, B., Hollander, K., Lefeber, D.: Compliant
  actuator designs.
\newblock IEEE Robotics \& Automation Magazine \textbf{16}(3), 81--94 (2009).
\newblock \urlprefix\url{http://ieeexplore.ieee.org/document/5233419/}

\bibitem{kopacek_robotized_2003}
Kopacek, P., Kopacek, B.: Robotized {{Disassembly}} of {{Mobile Phones}}.
\newblock IFAC Proc. Volumes \textbf{36}(23), 103--105 (2003).
\newblock
  \urlprefix\url{http://linkinghub.elsevier.com/retrieve/pii/S1474667017376693}

\bibitem{kopacek_intelligent_2006}
Kopacek, P., Kopacek, B.: Intelligent, flexible disassembly.
\newblock The International Journal of Advanced Manufacturing Technology
  \textbf{30}(5-6), 554--560 (2006).
\newblock \urlprefix\url{http://link.springer.com/10.1007/s00170-005-0042-9}

\bibitem{majewicz_design_2010}
Majewicz, A., Glasser, J., Bauer, R., Belkoff, S.M., Mears, S.C., Okamura,
  A.M.: Design of a haptic simulator for osteosynthesis screw insertion.
\newblock In: Haptics {Symposium}, 2010 {IEEE}, pp. 497--500. IEEE (2010)

\bibitem{liam}
Rujaevich, C., Lessard, J., Chandler, S., Shannon, S., Dahmus, J., Guzzo, R.:
  Liam - {An} {Innovation} {Story} (2016).
\newblock
  \urlprefix\url{https://www.apple.com/environment/pdf/Liam_white_paper_Sept2016.pdf}.
\newblock Bibtex: liam\_2016

\bibitem{tsetserukou2010}
Tsetserukou, D., Sato, K., Tachi, S.: Exointerfaces: Novel exosceleton haptic
  interfaces for virtual reality, augmented sport and rehabilitation.
\newblock In: Proceedings of the 1st Augmented Human International Conference,
  AH '10, pp. 1:1--1:6. ACM, New York, NY, USA (2010).
\newblock \doi{10.1145/1785455.1785456}.
\newblock \urlprefix\url{http://doi.acm.org/10.1145/1785455.1785456}

\bibitem{disassembly_book}
Vongbunyong, S., Chen, W.H.: Disassembly {{Automation}}.
\newblock Sustainable Production, Life Cycle Engineering and Management.
  {Springer International Publishing}, Cham (2015).
\newblock \doi{10.1007/978-3-319-15183-0}.
\newblock \urlprefix\url{http://link.springer.com/10.1007/978-3-319-15183-0}

\bibitem{vongbunyong_basic_2013}
Vongbunyong, S., Kara, S., Pagnucco, M.: Basic behaviour control of the
  vision‐based cognitive robotic disassembly automation.
\newblock Assembly Automation \textbf{33}(1), 38--56 (2013).
\newblock \doi{10.1108/01445151311294694}.
\newblock
  \urlprefix\url{http://www.emeraldinsight.com/doi/abs/10.1108/01445151311294694}

\end{thebibliography}


\pagebreak
\pagebreak

\end{document}